\documentclass[letterpaper, 10 pt, conference]{ieeeconf}  

\IEEEoverridecommandlockouts                              

\usepackage{pgfplots}
\usepackage{tikz}
\usepackage{epsfig}
\usepackage{graphicx}
\usepackage{amsmath}
\usepackage{amssymb}
\usepackage{algorithm2e}
\usepackage{algorithmic}
\usepackage{subcaption}
\usepackage{float}
\usepackage{booktabs}
\usepackage{balance}
\usepackage{cite}

\title{\LARGE \bf
Look and Listen: A Multi-modality Late Fusion \\Approach to Scene Classification for Autonomous Machines
}
\author{Jordan J. Bird$^{1,2}$, Diego R. Faria$^{1,2}$, Cristiano Premebida$^{3}$, Anik\'o Ek\'art$^{1}$ and George Vogiatzis$^{1,2}$ 
    \thanks{}
\thanks{$^{1}$J.J. Bird, D.R. Faria, A. Ekart, and G. Vogiatzis are with the School of Engineering and Applied Science,
        Aston University, Birmingham, United Kingdom. Emails: 
        {\tt\small \{birdj1, d.faria, a.ekart, g.vogiatzis\}@aston.ac.uk}}%
\thanks{$^{1}$J.J. Bird, D.R. Faria, and G. Vogiatzis are also with the Aston Robotics Vision and Intelligent Systems (ARVIS) Lab 
        {\tt\small https://arvis-lab.io/}}%
\thanks{$^{3}$C. Premebida is with the Institute of Systems and Robotics, Department of Electrical and Computer Engineering, University of Coimbra, Coimbra, Portugal. Email:
  {\tt\small cpremebida@isr.uc.pt}      }%
}

\begin{document}

\maketitle
\thispagestyle{empty}
\pagestyle{empty}

\begin{abstract}
The novelty of this study consists in a multi-modality approach to scene classification, where image and audio complement each other in a process of deep late fusion. The approach is demonstrated on a difficult classification problem, consisting of two synchronised and balanced datasets of 16,000 data objects, encompassing 4.4 hours of video of 8 environments with varying degrees of similarity. We first extract video frames and accompanying audio at one second intervals. The image and the audio datasets are first classified independently, using a fine-tuned VGG16 and an evolutionary optimised deep neural network, with accuracies of 89.27\% and 93.72\%, respectively. This is followed by late fusion of the two neural networks to enable a higher order function, leading to accuracy of 96.81\% in this multi-modality classifier with synchronised video frames and audio clips. The tertiary neural network implemented for late fusion outperforms classical state-of-the-art classifiers by around 3\% when the two primary networks are considered as feature generators. We show that situations where a single-modality may be confused by anomalous data points are now corrected through an emerging higher order integration. Prominent examples include a water feature in a city misclassified as a river by the audio classifier alone and a densely crowded street misclassified as a forest by the image classifier alone. Both are examples which are correctly classified by our multi-modality approach. 

\end{abstract}

\section{Introduction}
\textit{`Where am I?'} is a relatively simple question answered by human beings though it requires  exceptionally complex neural processes. Humans use their senses of vision, hearing, temperature etc. as well as past experiences to discern whether they happen to be indoors, outdoors, and geolocate in general. This process occurs, for all intents and purposes, in an instant. Visuo-auditory perception is optimally integrated by humans in order to solve ambiguities; it is widely recognised that audition dominates time perception while vision dominates space perception. Both modalities are essential for awareness of the surrounding environment~\cite{LLL_iccv_2017}. In a world rapidly moving towards autonomous machines outside of the laboratory or home, environmental recognition is an important piece of information which should be considered as part of interpretive processes of spatial awareness.

\begin{figure}[!t]
	\centering
	\includegraphics[width=1\columnwidth, trim={2.7cm 9.2cm 8.2cm 1.9cm},clip]{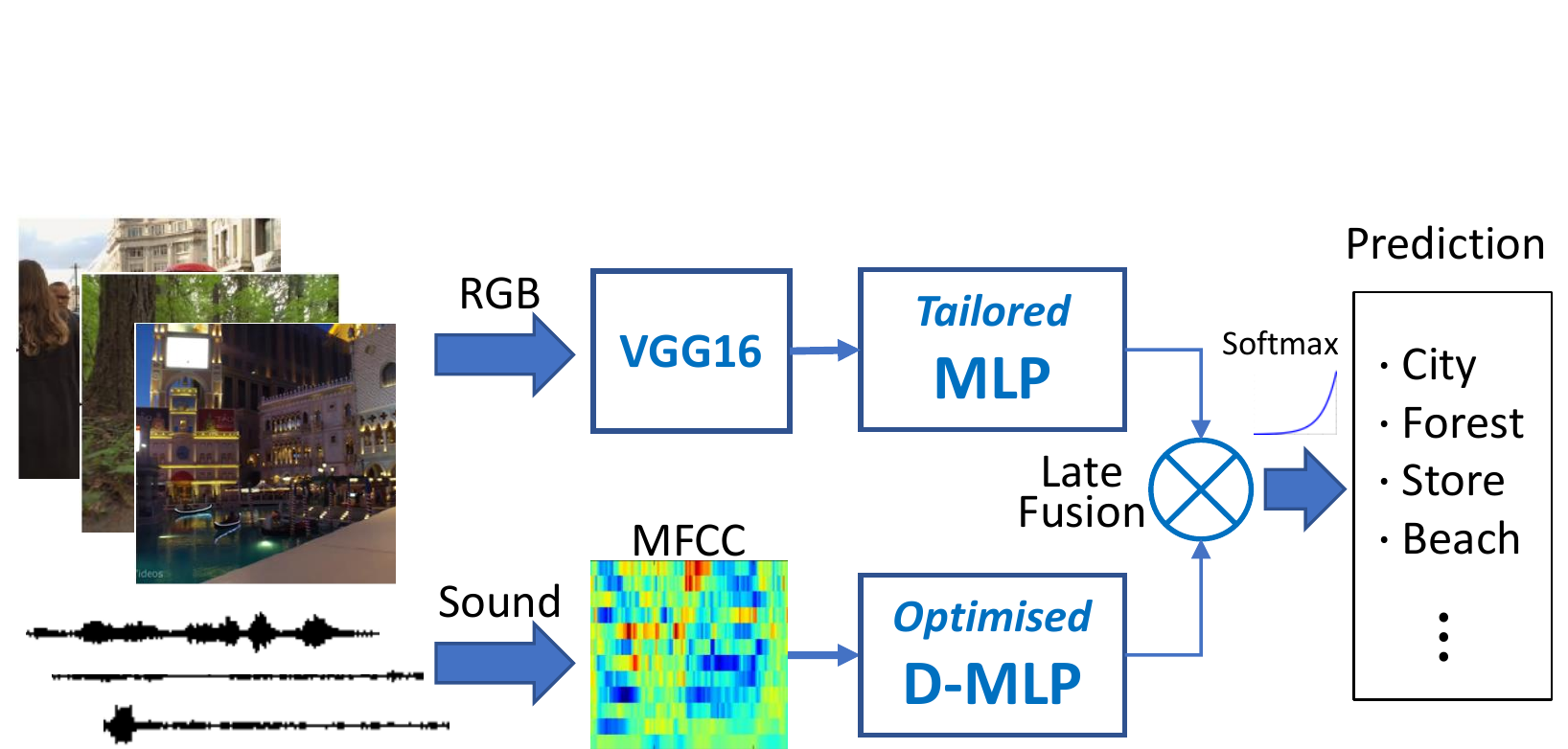}
	\caption{The proposed multi-modality (video and audio) approach to scene classification. MFCC are extracted from audio-frames as input to an optimised DNN, while VGG16 and an ANN classify the images. We propose a higher-order function to perform late fusion.}
	\label{fig:overview}
\end{figure}


\begin{figure*}[!t]
    \centering
    \includegraphics[scale=0.75]{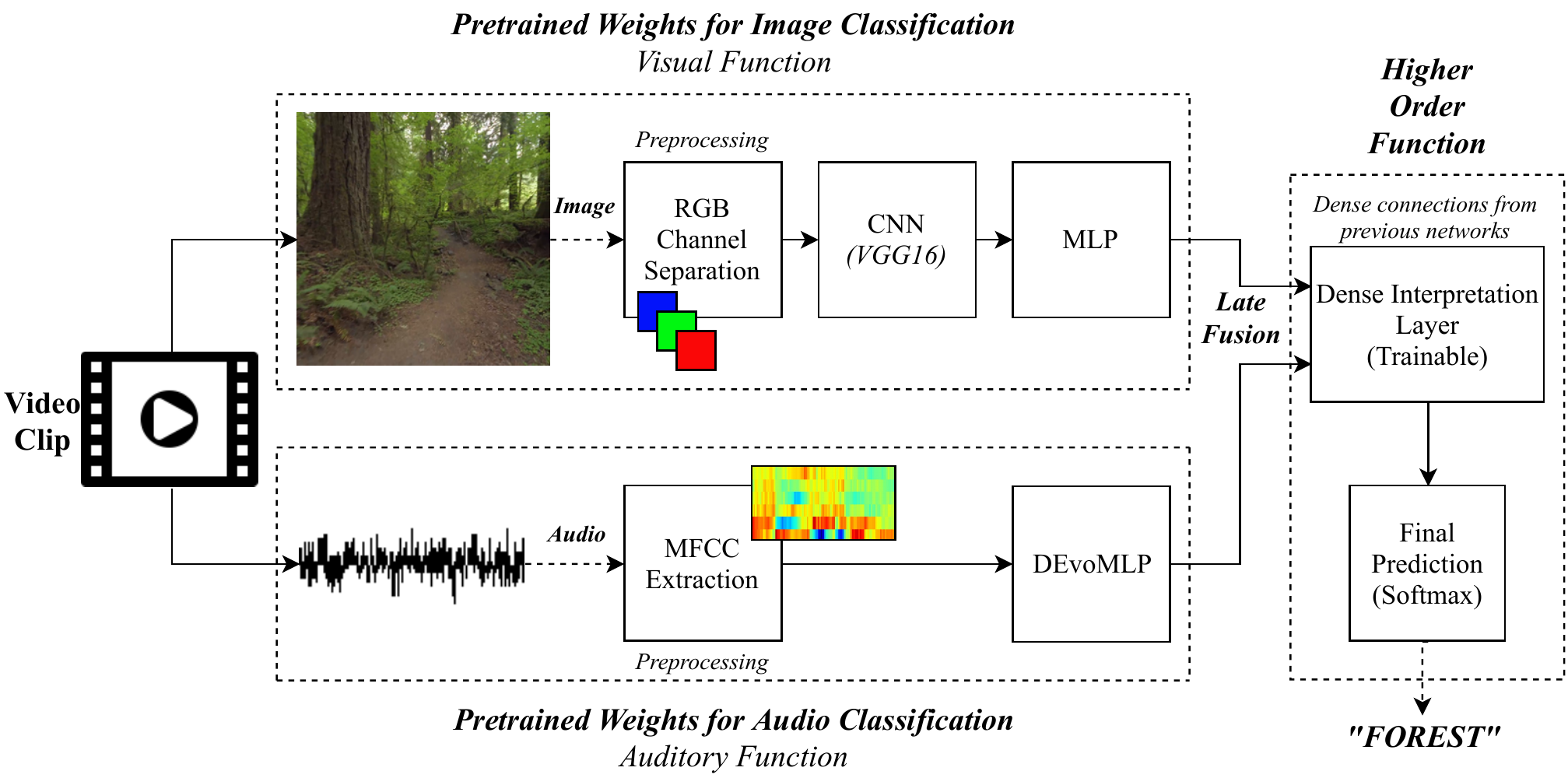}
    \caption{Overview of the multi-modality network. Pre-trained networks without softmax activation layer take synchronised images and audio segments as input, and classify based on interpretations of the outputs of the two models.}
    \label{fig:modeloverview}
\end{figure*}

Current trends in Robotic Vision~\cite{Geiger_pami,Ufranke_scen,ground_recog,onda2018dynamic}  indicate two main reasons for the usefulness of scene classification. The most obvious reason is simply the ability of an awareness of where one currently is, but furthermore, and in more complex situations, the awareness of one's surroundings can be further used as input to learning models or as a parameter within an intelligent decision making process. 
Just as humans \textit{'classify'} their surroundings for every day navigation and reasoning, this ability will very soon become paramount for the growing field of autonomous machines in the outside world such as self-driving cars and self-flying drones, and possibly, autonomous humanoid androids further into the future. Related work (Section \ref{sec:relatedwork}) finds that although the processes of classification themselves are well-explored, multi-modality classification is a ripe area enabled by the rapidly increasing hardware limits faced by researchers and consumers. With this finding in mind, we explore a bi-modal sensory cue combination for environment recognition as illustrated in Figure \ref{fig:overview}. This endows the autonomous machine with the ability to \textit{look} (Computer Vision) and to \textit{hear} (Audio Processing) before predicting the environment with a late fusion interpretation network for higher order functions such as anomaly detection and decision making. The main motivation for this is to disambiguate the classification process; for example, if a person were to observe busy traffic on a country road, the sound of the surroundings alone could be misclassified as a city street, whereas vision enables the observer to recognise the countryside and correct this mistake. Conversely, a densely crowded city street confuses a strong vision model since no discernable objects are recognised at multiple scales, but the sounds of the city street can still be heard. Though this anomalous data point has confused the visual model, the interpretation network learns these patterns, and the audio classification is given precedence leading to a correct prediction. The main contributions of this study are centred around the proposed multi-modality framework illustrated in Figure~\ref{fig:modeloverview} and are the following: (1)~A large dataset encompassing multiple dynamic environments is formed and made publicly available.\footnote{Full dataset is available at:\\ https://www.kaggle.com/birdy654/scene-classification-images-and-audio} This dataset provides a challenging problem, since many environments have similar visual and audio features. 
(2)~Supervised Transfer Learning of the VGG16 model towards scene classification by training upon the visual data, together with engineering a range of interpretation neurons for fine-tuning, lead to accurate classification abilities. 
(3)~The evolutionary optimisation of a deep neural network for audio processing of attributes extracted from the accompanying audio leads to accurate classification abilities, similarly to the vision network.
(4)~The final late fusion model combines and interprets the output of previously trained networks in order to discern and correct various anomalous data points that led to mistakes (examples of this are given in Section \ref{subsec:comparisonandexploration}). The multi-modality model outperforms both the visual and audio networks alone, therefore we argue that multi-modality classification is a better solution for scene classification.

\section{Related Work}
\label{sec:relatedwork}
Much state-of-the-art work in scene classification explores the field of autonomous navigation in self-driving cars. Many notable recent studies~\cite{Lidar_city,robotic_Zach,Lidar_fusion} find dynamic environment mapping leading to successful real-time navigation and object detection through LiDAR data. Visual data in the form of images are often shown to be useful in order to observe and classify an environment; notably 66.2\% accuracy was achieved on a large scene dataset through transfer learning from the Places CNN\footnote{http://places.csail.mit.edu/downloadCNN.html} compared to ImageNet transfer learning and SVM which achieved only 49.6\%~\cite{zhou2014learning}. Similarly Xie, et al.~\cite{xie2015hybrid} found that through a hybrid CNN trained for scene classification, scores of 82.24\% were achieved for the ImageNet dataset.\footnote{http://www.image-net.org} Though beyond the current capabilities of autonomous machine hardware, an argument has recently been put forward for temporal awareness through LSTM~\cite{byeon2015scene}, achieving 78.56\% and 70.11\% pixel accuracy on two large image datasets. A previous single-modality study found improvement of scene classification ability by transferring from both VGG16 and scene images from videogames to photographic images of real-life environments~\cite{bird2020cnnsimulation} with an average improvement of +7.15\% when simulation data was present prior to transfer of weights. 
In terms of audio, the usefulness of MFCC audio features in statistical learning for recognition of environment has recently been shown~\cite{chu2006scene}, gaining classification accuracies of 89.5\%, 89.5\% and 95.1\% with KNN, GMM, and SVM methods respectively.  Nearest-neighbour MFCC classification of 25 environments achieved 68.4\% accuracy compared to a subject group of human beings who on average recognised environments from audio data with 70\% accuracy \cite{peltonen2002computational}. It is argued that a deep neural network outperforms an SVM for scene classification from audio data, gaining up to 92\% accuracy ~\cite{petetin2015deep}.

Researchers have shown that human beings use multiple parts of the brain for general recognition tasks, including the ability of environmental awareness~\cite{mattson2014superior,eysenck2015cognitive}. Though in many of these studies a single-modality is successful, we argue that, since the human brain merges the senses into a robust percept for recognition tasks, the field of scene classification should find some loose inspiration from this process through data fusion. We explore visual and audio in this experiment due to accessibility, since there is a lot of audio-visual video data available to researchers. We propose that in the future  further sensory data are explored, given the success of this preliminary experiment (Section \ref{sec:futureconclusions}). 

\section{Problem and Method}
\label{sec:method}
The state-of-the-art is to interpret real-world data considering a single input. Conversely, the idea of multi-modality learning is to consider multiple forms of input~\cite{srivastava2012multimodal}. 

\begin{figure*}[t!]
    \centering
    \includegraphics[scale=0.6]{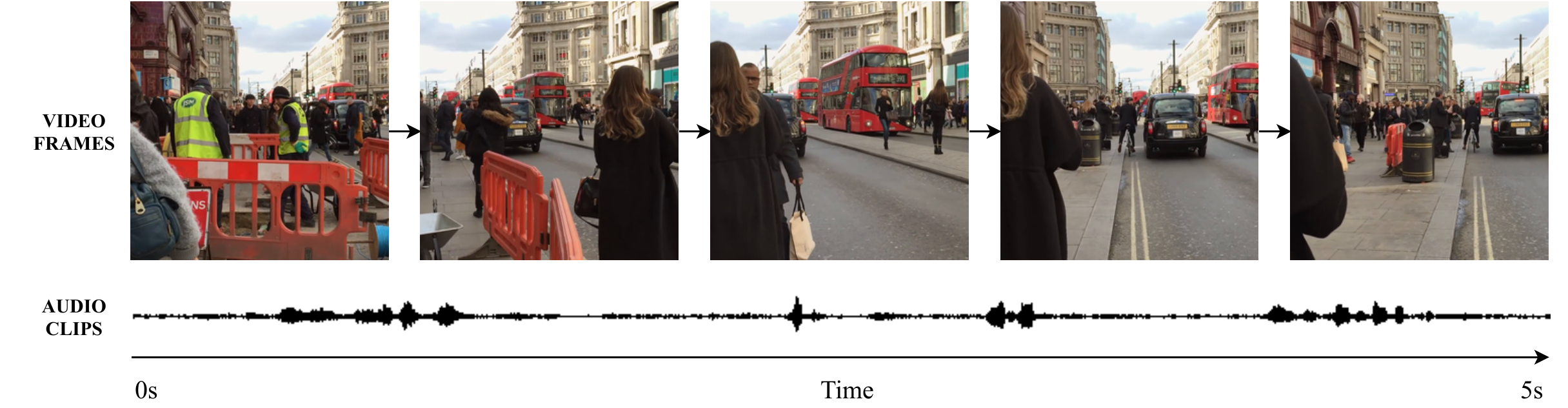}
    \caption{Example of extracted data from a five second timeline. Each second, a frame is extracted from the video along with the accompanying second of audio.}
    \label{fig:timeline}
\end{figure*}

Simply put, the question posed to a classifier is \textit{`where are you?'}. Synchronised images and audio are treated as inputs to the classifier, and are labelled semantically. A diagram of this process can be observed in Figure \ref{fig:modeloverview};\footnote{VGG Convolutional Topology is detailed in \cite{Simonyan14c}} visual and auditory functions consider synchronised image and audio independently, before a higher order function occurs. The two neural networks are concatenated into an interpretation network via late fusion to a further hidden layer before a final prediction is made.
Following dataset acquisition of videos, video frames and accompanying audio clips, the general experimental processes are as follows. (i) \textit{For audio classification:} the extraction of MFCCs of each audio clip to generate numerical features and evolutionary optimisation of neural network topology to derive network hyperparameters. (ii) \textit{For image classification:} pre-processing through a centre-crop (square) and resizing to a 128x128x3 RGB matrix due to the computational complexity required for larger images, and subsequent fine tuning of the interpretation layers for fine-tune transfer learning of the VGG16 trained weight set. (iii) \textit{For the final model:} freeze the trained weights of the first two models while benchmarking an interpretation layer for synchronised classification of both visual and audio data. This process is described in more detail throughout Subsections \ref{subsec:dataset} and \ref{subsec:machinelearning}. 

\subsection{Dataset Acquisition}
\label{subsec:dataset}
Initially, 45 videos as sources are collected in varying length for 9 environmental classes at NTSC 29.97 FPS and are later reduced to 2000 seconds each: \textit{Beach} (4 sources, 2080 seconds), \textit{City} (5 sources, 2432 seconds), \textit{Forest} (3 sources, 2000 seconds), \textit{River} (8 sources, 2500 seconds) \textit{Jungle} (3 sources, 2000 seconds), \textit{Football Match} (4 sources, 2300 seconds), \textit{Classroom} (6 sources, 2753 seconds), \textit{Restaurant} (8 sources, 2300 seconds), and \textit{Grocery Store} (4 sources, 2079 seconds). 
The videos are dynamic, from the point of view of a human being. All audio is naturally occurring within the environment. It must be noted that some classes are similar environments and thus provide a difficult recognition problem. To generate the initial data objects, a crop is performed at each second. The central frame of the second of video is extracted with the accompanying second of audio, an example of data processing for a city is shown in Figure \ref{fig:timeline}. Further observation lengths should be explored in future. 
This led to 32,000 data objects, 16,000 images (128x128x3 RGB matrices) accompanied by 16,000 seconds (4.4 hours) of audio data. 
We then extract the the Mel-Frequency Cepstral Coefficients (MFCC)~\cite{muda2010voice} of the audio clips through a set of sliding windows 0.25s in length (ie frame size of 4K sampling points) and an additional set of overlapping windows, thus producing 8 sliding windows. From each audio frame, we extract 13 MFCC attributes, producing 104 attributes per 1 second clip. MFCC extraction consists of the following steps: The Fourier Transform (FT) of the time window data $\omega$ is derived as $X(j\omega)=\int_{-\infty}^\infty x(t) e^{-j\omega t} dt$. The powers from the FT are mapped to the Mel scale, the psychological scale of audible pitch~\cite{stevens1937scale}. This occurs through the use of a triangular temporal window. The Mel-Frequency Cepstrum (MFC), or power spectrum of sound, is considered and logs of each of their powers are taken. The derived Mel-log powers are treated as a signal, and a Discrete Cosine Transform (DCT) is measured. This is given as $X_k =  \sum_{n=0}^{N-1} x_n  cos  \begin{bmatrix}  \frac{\pi}{N} (n+ \frac{1}{2}) k \end{bmatrix}$ where $k=0,...,N-1$ is the index of the output coefficient being calculated and $x$ is the array of length $N$ being transformed.
The amplitudes of the spectrum are known as the MFCCs.

The learning process we present is applicable to consumer-level hardware (unlike temporal techniques) and thus accessible for the current abilities of autonomous machines.

\subsection{Machine Learning Processes}
\label{subsec:machinelearning}
For audio classification, an evolutionary algorithm~\cite{bird2019evolutionary} was used to select the amount of layers and neurons contained within a MLP in order to derive the best network topology. Population is set to 20 and generations to 10, since stabilisation occurs prior to generation 10. The simulation is executed five times in order to avoid stagnation at local minima being taken forward as a false best solution. Activations of the hidden layers are set to ReLu. For image classification, the VGG16 layers and weights~\cite{Simonyan14c} are implemented except the dense interpretation layers beyond the Convolutional layers, which is then followed by $\{2, 4, 8, \cdots, 4096\}$ ReLu neurons for interpretation and finally a softmax activated layer towards the nine-class problem. In order to generate the final model, the previous process of neuron benchmarking is also followed. The two trained models for audio and image classification have their weights frozen, and training concentrates on the interpretation of the outputs of the networks. Referring back to Figure \ref{fig:modeloverview}, the softmax activation layers are removed from the initial two networks in order to pass their interpretations to the final interpretation layer through concatenation, a densely connected layer following the two networks and $\{2, 4, 8, \cdots, 4096\}$ ReLu neurons are benchmarked in order to show multi-modality classification ability. All neural networks are trained for 100 epochs with shuffled 10-fold cross-validation.

\section{Experimental Results}
\label{sec:results}
\subsection{Fine Tuning of VGG16 Weights and Topology}
\begin{figure}
\centering
\begin{tikzpicture}[scale=0.75]
\footnotesize
  \begin{axis}[
    ybar,
    legend style={at={(0.5,-0.4)},
      anchor=north,legend columns=-1},
    ylabel={Image Classification Accuracy (\%)},
    symbolic x coords={2,4,8,16,32,64,128,256,512,1024,2048,4096},
    xtick=data,
    ymin=10, ymax=100,
    ytick={10,20,30,40,50,60,70,80,90,100},
    x tick label style={rotate=45,anchor=east},
    xlabel={Interpretation Neurons},
    grid style=dashed,
    ]
    \addplot coordinates {(2, 15.22) (4, 14.08) (8, 25.25) (16, 27.47) (32, 43.22) (64, 51.84) (128, 38.25) (256, 53.79) (512, 72.89) (1024, 82.82) (2048, 89.28) (4096, 77.4)};
		
  \end{axis}
\end{tikzpicture}
\caption{Image 10-fold Classification Accuracy corresponding to interpretation neuron numbers.}
\label{fig:vgg16graph}
\end{figure}
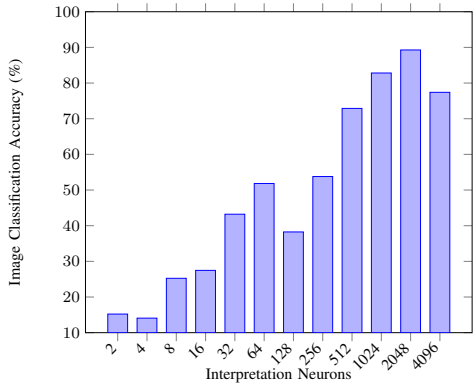

Figure \ref{fig:vgg16graph} shows the tuning of interpretation neurons for the image classification network. The best result was 89.27\% 10-fold classification accuracy, for 2048 neurons. 

\subsection{Evolving the Sound Processing Network}
\begin{table}[]
\centering
\caption{Final results of the (\#) five Evolutionary Searches sorted by 10-fold validation Accuracy. \textit{Conns.} denotes the number of connections in the network.}
\label{tab:evoresults}
\footnotesize
\begin{tabular}{@{}lllr@{}}
\toprule
\textbf{Simulation} & \textbf{Hidden Neurons} & \textbf{Connections} & \multicolumn{1}{l}{\textbf{Accuracy}} \\ \midrule
\textit{\textbf{2}} & 977, 365, 703, 41       & 743,959         & \textbf{93.72\%}                      \\
\textit{\textbf{4}} & 1521, 76, 422, 835      & 664,902         & 93.54\%                               \\
\textit{\textbf{1}} & 934, 594, 474           & 937,280         & 93.47\%                               \\
\textit{\textbf{3}} & 998, 276, 526, 797, 873 & 1,646,563       & 93.45\%                               \\
\textit{\textbf{5}} & 1524, 1391, 212, 1632   & 2,932,312       & 93.12\%                               \\ \bottomrule
\end{tabular}
\end{table}
Regardless of initial (random) population, stabilisation of the audio network topology search occurred around the 92-94\% accuracy mark. The best solution was a deep network of \textit{977, 365, 703, 41} hidden-layer neurons, which gained 93.72\% accuracy via 10-fold cross validation. All final solutions are presented in Table \ref{tab:evoresults}. Interestingly, a less complex solution scores a competitive score of 93.54\% accuracy with 79,057 fewer network connections.

\subsection{Fine Tuning the Final Model}
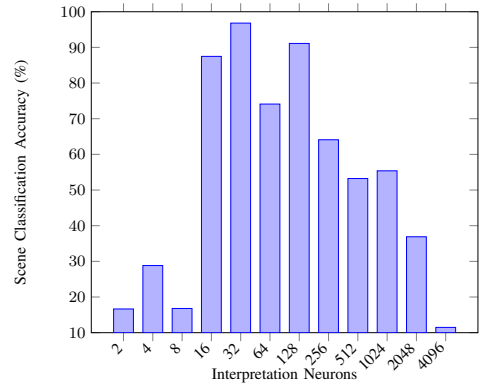
\begin{figure}
\centering
\begin{tikzpicture}[scale=0.75]
\footnotesize
  \begin{axis}[
    ybar,
    legend style={at={(0.5,-0.4)},
      anchor=north,legend columns=-1},
    ylabel={Scene Classification Accuracy (\%)},
    symbolic x coords={2,4,8,16,32,64,128,256,512,1024,2048,4096},
    xtick=data,
    ymin=10, ymax=100,
    ytick={10,20,30,40,50,60,70,80,90,100},
    x tick label style={rotate=45,anchor=east},
    xlabel={Interpretation Neurons},
    grid style=dashed,
    ]
    \addplot coordinates {(2, 16.65) (4, 28.83) (8, 16.79) (16, 87.48) (32, 96.81) (64, 74.11) (128, 91.11) (256, 64.1) (512, 53.23) (1024, 55.39) (2048, 36.9) (4096, 11.48)};
		
  \end{axis}
\end{tikzpicture}
\caption{Multi-modality 10-fold Classification Accuracy corresponding to interpretation neuron numbers.}
\label{fig:multigraph}
\end{figure}

With the two input networks frozen at the previously trained weights, the results of the multi-modality network can be observed in Figure \ref{fig:multigraph}. The best interpretation layer was selected as 32, which attained a classification accuracy of 96.81\% as shown in Table \ref{tab:multigraph}. Late fusion was tested with other models by treating the two networks as feature generators for input, a Random Forest scored 94.21\%, Naive Bayes scored 93.61\% and an SVM scored 95.08\%, which were all outperformed by the tertiary deep neural network.

\begin{figure}
    \centering
    \includegraphics[scale=0.45]{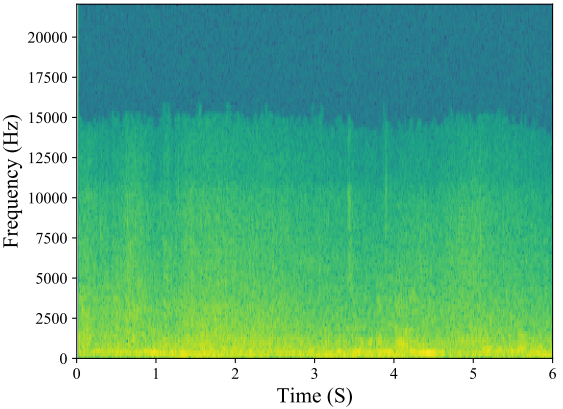}
    \caption{Beach sonogram (with speech at 3s-4.5s)}
    \label{fig:beachsono}
\end{figure}

\begin{figure}
    \centering
    \includegraphics[scale=0.45]{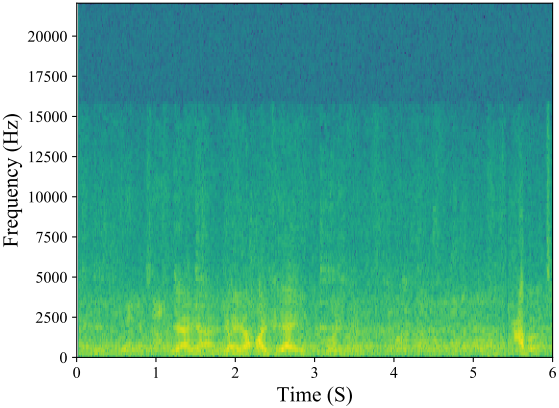}
    \caption{Restaurant sonogram (with speech throughout)}
    \label{fig:restaurantsono}
\end{figure}

\begin{figure}[]
    \centering
    \includegraphics[scale=0.75]{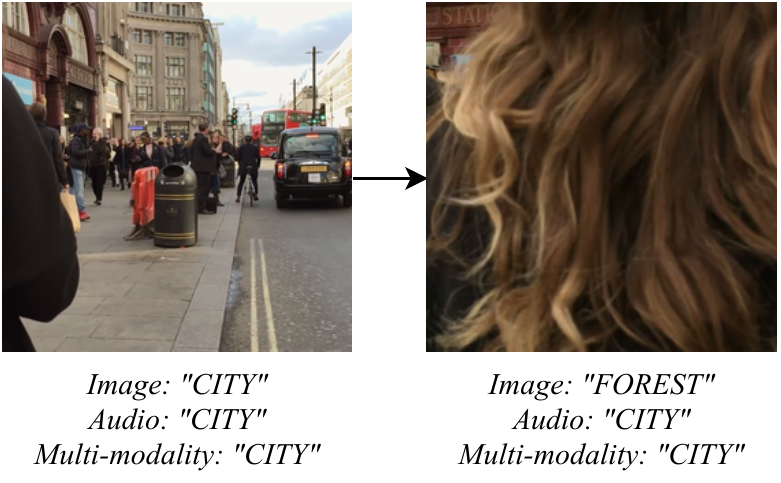}
    \caption{An example of confusion of the vision model, which is corrected through multi-modality. In the second frame, the image of hair is incorrectly classified as the ``FOREST" environment through Computer Vision.}
    \label{fig:confusion}
\end{figure}

\begin{figure}
    \centering
    \includegraphics[scale=0.75]{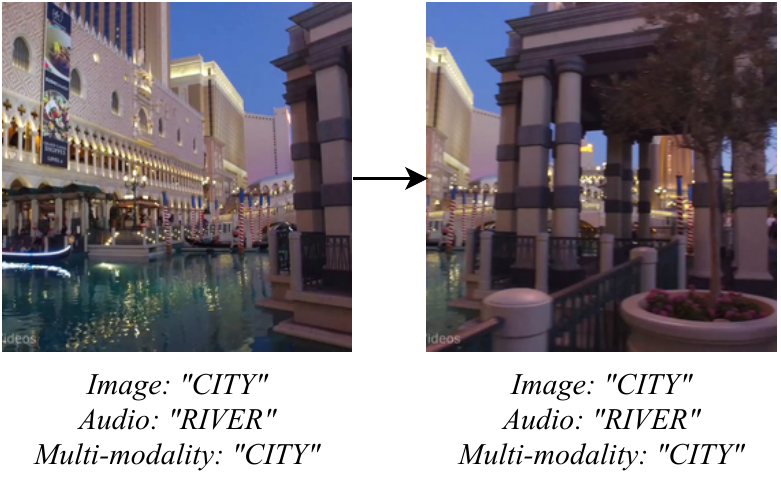}
    \caption{An example of confusion of the audio model, which is corrected through multi-modality. In both examples, the audio of a City is incorrectly classified as the ``RIVER" environment due to the sounds of a fountain and flowing water by the audio classification network.}
    \label{fig:confusion2}
\end{figure}

\subsection{Comparison and Analysis of Models}
\label{subsec:comparisonandexploration}
\begin{table}[t]
\centering
\caption{Scene Classification ability of the three tuned models on the dataset}
\label{tab:multigraph}
\footnotesize
\begin{tabular}{@{}lr@{}}
\toprule
\textbf{Model}                & \multicolumn{1}{l}{\textbf{Scene Classification Ability}} \\ \midrule
\textit{\textbf{Visual}}      & 89.27\%                                                      \\
\textit{\textbf{Auditory}}    & 93.72\%                                                      \\
\textit{\textbf{Multi-modality}} & \textbf{96.81\%}                                             \\ \bottomrule
\end{tabular}
\end{table}

For final comparison of classification models, Table \ref{tab:multigraph} shows the best performances of the tuned vision, audio, and multi-modality models, through 10-fold cross validation. Though visual classification was the most difficult task at 89.27\% prediction accuracy, it was only slightly outperformed by the audio classification task at 93.72\%. Outperforming both models was the multi-modality approach (Figure \ref{fig:modeloverview}), when both vision and audio are considered through network concatenation, the model learns not only to classify both network outputs concurrently, but more importantly calculates relationships between them. An example of confusion of the audio model can be seen by the sonograms in Figures \ref{fig:beachsono} and \ref{fig:restaurantsono}. Multiple frames of audio from the beach clip were mis-classified as \textit{`Restaurant'} due to focus on the human speech audio. The image classification model on the other hand correctly classified these frames, and the multi-modality model did also. The same was observed several times within the classes \textit{`City'}, \textit{`Grocery Store'}, and \textit{`Football Match'}. We note that these Sonograms show the frequency of the raw audio (stereo averaged into mono) for demonstrative purposes and MFCC extraction occurs after this point. Another example of this can be seen in Figure \ref{fig:confusion}, in which the Vision model has been confused by a passerby. The audio model recognises the sounds of traffic and crowds etc. (this is also possibly why the audio model outperforms the image model slightly), the interpretation network has learnt this pattern and thus has \textit{`preferred'} the outputs of the audio model in this case. Since the multi-modality model outperforms both single-modality models, this confusion also occurs in the opposite direction; observe that in Figure \ref{fig:confusion2}, the audio model has inadvertently predicted that the environment is a river due to the sounds of water, yet the image classifier correctly predicts that it is a city, in this case, Las Vegas. The multi-modality model, again, has learnt such patterns and has preferred the prediction of the image model, leading to a correct recognition of environment.

\begin{table}[]
\caption{Results of the three approaches applied to completely unseen data (9 classes)}
\label{tab:unseendata}
\footnotesize
\begin{tabular}{@{}lll@{}}
\toprule
\textbf{Approach}                      & \textbf{Correct/Incorrect} & \textbf{Classification Accuracy} \\ \midrule
\textit{\textbf{Audio Classification}} & 359/1071                   & 33.52\%                          \\
\textit{\textbf{Image Classification}} & 706/1071                   & 65.92\%                          \\
\textit{\textbf{Multi-modality}}       & \textbf{856/1071}          & \textbf{79.93\%}                 \\ \bottomrule
\end{tabular}
\end{table}

\begin{figure}
    \centering
    \includegraphics[scale=0.75]{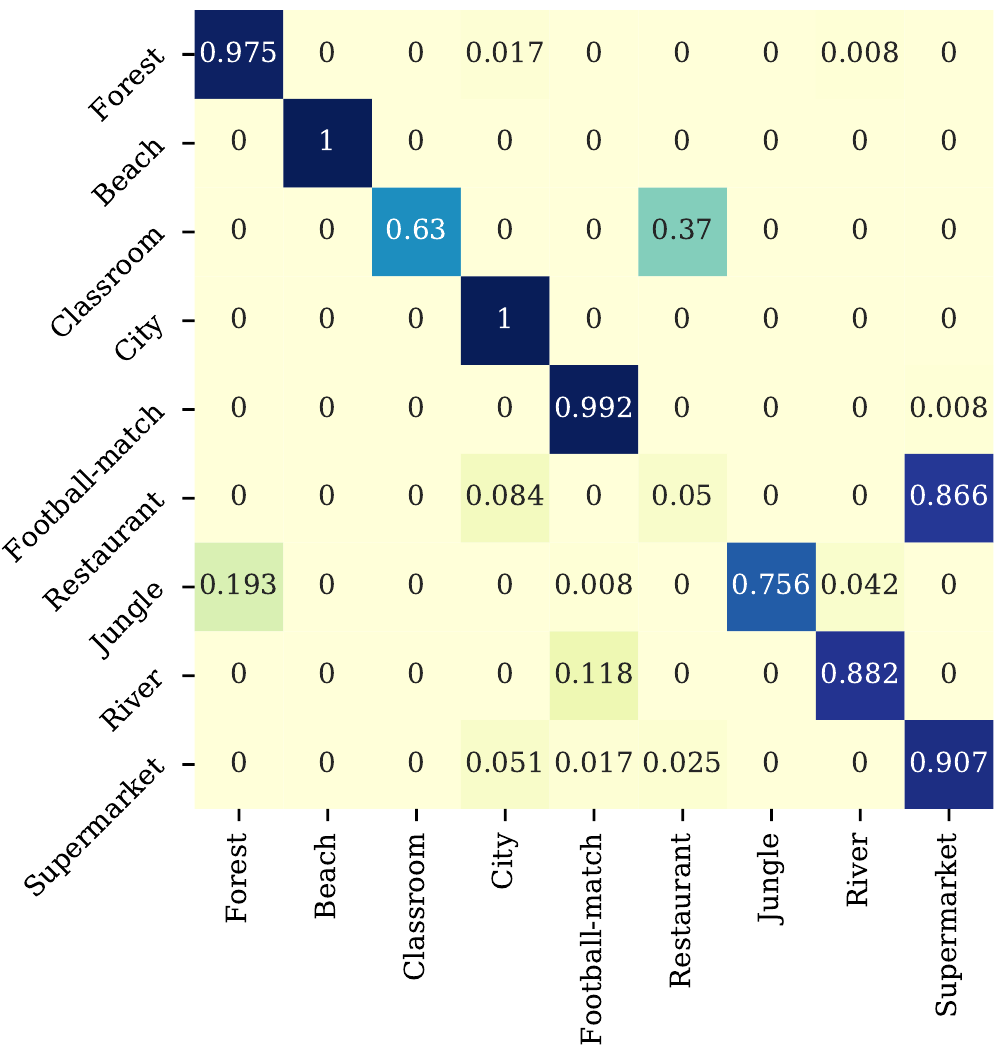}
    \caption{Confusion matrix for the multi-modality model applied to completely unseen data}
    \label{fig:errormatrix}
\end{figure}
The results of applying the models to completely unseen data (two minutes per class) can be seen in Table \ref{tab:unseendata}. It can be observed that audio classification of environments is weak at 33.52\%, which is outperformed by image classification at 65.92\% accuracy. Both approaches are outperformed by the multi-modality approach which scores 79.93\% classification accuracy. The confusion matrix of the multi-modality model can be observed in Figure \ref{fig:errormatrix}; the main issue is caused by 'Restaurant' being confused as `Supermarket', while all other environments are classified strongly. On manual observation, both classes in the unseen data both feature a large number of people with speech audio, we conjecture that this is possibly most similar to the supermarkets in the training dataset and thus the model is confident that both of these classes belongs to supermarket. This suggests that the data could be more diversified in future in order to feature more minute details and thus improve the model's abilities for discerning between the two.
\section{Conclusions and Future Work}
\label{sec:futureconclusions}
This study presented and analysed three scene classification models: (a) a vision model through fine-tuned VGG16 weights for classification of images of environments. (b) a deep neural network for classification of audio of environments and (c), a multi-modality approach, which outperformed the two original approaches through the gained ability of detection of anomalous data through consideration of the outputs of both models. The tertiary neural network for late fusion was compared and found to be superior to Naive Bayes, Random Forest, and Support Vector Machine classifiers. We argue that since audio classification is a relatively easy task, it should be implemented where available to improve environmental recognition tasks. This work focused on the context of autonomous machines, and thus consumer hardware capability was taken into account through temporal-awareness implemented within the feature extraction process rather than within the learning process. In future, better results could be gained from attempting to enable a neural network to learn temporal awareness in recurrence. Since the model was found to be effective with the complex problem posed through our dataset, future studies could concern other publicly available datasets in order to explore the applicability more widely. With the available hardware, evolutionary selection of network topology was only possible with the audio classifier. In future and with more resources, this algorithm could be applied to both the vision and interpretation models with the expectation to achieve a better set of hyperparameters beyond the tuning performed in this study. The model could also be applied in real-world scenarios. For example, it  has recently been shown that autonomous environment detection is useful in the automatic application of scene settings for hearing aids~\cite{buchler2005sound}. Future works could also consider optimisation of the frame segmentation process itself as well as exploration of the possibility of multiple image inputs per task. Additionally, given the success of late fusion in this work, applications to video classification tasks could be considered through a similar approach. 

\small
\bibliography{egbib}

\begin{thebibliography}{10}

\bibitem{LLL_iccv_2017}
R.~{Arandjelovic} and A.~{Zisserman}, ``Look, listen and learn,'' in {\em IEEE
  International Conference on Computer Vision (ICCV)}, pp.~609--617, 2017.

\bibitem{Geiger_pami}
A.~{Geiger}, M.~{Lauer}, C.~{Wojek}, C.~{Stiller}, and R.~{Urtasun}, ``3{D}
  traffic scene understanding from movable platforms,'' {\em IEEE Transactions
  on Pattern Analysis and Machine Intelligence (PAMI)}, vol.~36, no.~5,
  pp.~1012--1025, 2014.

\bibitem{Ufranke_scen}
M.~{Cordts}, T.~{Rehfeld}, M.~{Enzweiler}, U.~{Franke}, and S.~{Roth},
  ``Tree-structured models for efficient multi-cue scene labeling,'' {\em IEEE
  TPAMI}, vol.~39, no.~7, pp.~1444--1454, 2017.

\bibitem{ground_recog}
J.~{Xue}, H.~{Zhang}, and K.~{Dana}, ``Deep texture manifold for ground terrain
  recognition,'' in {\em IEEE/CVF CVPR}, pp.~558--567, 2018.

\bibitem{onda2018dynamic}
K.~Onda, T.~Oishi, and Y.~Kuroda, ``Dynamic environment recognition for
  autonomous navigation with wide {FOV} 3{D}-{L}i{DAR},'' {\em
  IFAC-PapersOnLine}, vol.~51, no.~22, pp.~530--535, 2018.

\bibitem{Lidar_city}
F.~{Yu}, J.~{Xiao}, and T.~{Funkhouser}, ``Semantic alignment of {L}i{DAR} data
  at city scale,'' in {\em IEEE Conference on Computer Vision and Pattern
  Recognition (CVPR)}, pp.~1722--1731, 2015.

\bibitem{robotic_Zach}
C.~{Zach}, A.~{Penate-Sanchez}, and M.~{Pham}, ``A dynamic programming approach
  for fast and robust object pose recognition from range images,'' in {\em IEEE
  CVPR}, pp.~196--203, 2015.

\bibitem{Lidar_fusion}
D.~{Xu}, D.~{Anguelov}, and A.~{Jain}, ``Point{F}usion: Deep sensor fusion for
  3{D} bounding box estimation,'' in {\em IEEE/CVF CVPR}, pp.~244--253, 2018.

\bibitem{zhou2014learning}
B.~Zhou, A.~Lapedriza, J.~Xiao, A.~Torralba, and A.~Oliva, ``Learning deep
  features for scene recognition using places database,'' in {\em Advances in
  neural information processing systems (NIPS)}, pp.~487--495, 2014.

\bibitem{xie2015hybrid}
G.-S. Xie, X.-Y. Zhang, S.~Yan, and C.-L. Liu, ``Hybrid cnn and
  dictionary-based models for scene recognition and domain adaptation,'' {\em
  IEEE Transactions on Circuits and Systems for Video Technology}, vol.~27,
  no.~6, pp.~1263--1274, 2015.

\bibitem{byeon2015scene}
W.~Byeon, T.~M. Breuel, F.~Raue, and M.~Liwicki, ``Scene labeling with lstm
  recurrent neural networks,'' in {\em IEEE Conf. on Computer Vision and
  Pattern Recognition (CVPR)}, pp.~3547--3555, 2015.

\bibitem{bird2020cnnsimulation}
J.~J. Bird, D.~R. Faria, P.~P. Ayrosa, and A.~Ek{\'a}rt, ``From simulation to
  reality: Cnn transfer learning for scene classification,'' in {\em 2020
  International Conference on Intelligent Systems (IS)}, IEEE, 2020.

\bibitem{chu2006scene}
S.~Chu, S.~Narayanan, C.-C.~J. Kuo, and M.~J. Mataric, ``Where am {I}? {S}cene
  recognition for mobile robots using audio features,'' in {\em IEEE
  International conference on multimedia and expo}, pp.~885--888, 2006.

\bibitem{peltonen2002computational}
V.~Peltonen, J.~Tuomi, A.~Klapuri, J.~Huopaniemi, and T.~Sorsa, ``Computational
  auditory scene recognition,'' in {\em IEEE International Conference on
  Acoustics, Speech, and Signal Processing}, vol.~2, pp.~II--1941, 2002.

\bibitem{petetin2015deep}
Y.~Petetin, C.~Laroche, and A.~Mayoue, ``Deep neural networks for audio scene
  recognition,'' in {\em IEEE 23rd European Signal Processing Conference
  (EUSIPCO)}, pp.~125--129, 2015.

\bibitem{mattson2014superior}
M.~P. Mattson, ``Superior pattern processing is the essence of the evolved
  human brain,'' {\em Frontiers in neuroscience}, vol.~8, p.~265, 2014.

\bibitem{eysenck2015cognitive}
M.~W. Eysenck and M.~T. Keane, {\em Cognitive psychology: A student's
  handbook}.
\newblock Psychology press, 2015.

\bibitem{srivastava2012multimodal}
N.~Srivastava and R.~Salakhutdinov, ``Multimodal learning with deep {B}oltzmann
  machines,'' in {\em Advances in Neural Information Processing Systems
  (NIPS)}, (USA), pp.~2222--2230, 2012.

\bibitem{Simonyan14c}
K.~Simonyan and A.~Zisserman, ``Very deep convolutional networks for
  large-scale image recognition,'' in {\em International Conference on Learning
  Representations}, 2015.

\bibitem{muda2010voice}
L.~Muda, M.~Begam, and I.~Elamvazuthi, ``Voice recognition algorithms using mel
  frequency cepstral coefficient ({MFCC}) and dynamic time warping ({DTW})
  techniques,'' {\em arXiv preprint arXiv:1003.4083}, 2010.

\bibitem{stevens1937scale}
S.~S. Stevens, J.~Volkmann, and E.~B. Newman, ``A scale for the measurement of
  the psychological magnitude pitch,'' {\em The Journal of the Acoustical
  Society of America}, vol.~8, no.~3, pp.~185--190, 1937.

\bibitem{bird2019evolutionary}
J.~J. Bird, A.~Ek{\'a}rt, C.~D. Buckingham, and D.~R. Faria, ``Evolutionary
  optimisation of fully connected artificial neural network topology,'' in {\em
  Intelligent Computing-Proceedings of the Computing Conference}, pp.~751--762,
  Springer, 2019.

\bibitem{buchler2005sound}
M.~B{\"u}chler, S.~Allegro, S.~Launer, and N.~Dillier, ``Sound classification
  in hearing aids inspired by auditory scene analysis,'' {\em Journal on
  Advances in Signal Processing}, no.~18, pp.~387--845, 2005.

\end{thebibliography}
\bibliographystyle{ieeetr}


\end{document}